\documentclass{templates/tinamemo}
\usepackage[dvips]{graphicx,rotating}
\usepackage{subfigure}
\usepackage[]{amsmath, amssymb, mathrsfs}
\usepackage{multirow}
\usepackage{multicol}
\usepackage{algorithm}

\begin{document}
\thispagestyle{empty}
Tina Memo No. 2018-004 \\               
Green Open Access pre-print of publication in Bioinformatics 2018 Mar 14. doi: 10.1093.

See also 2017-006 and 2015-014.

\vspace{3cm}

\begin{center}
{\huge
A New Method for the High-Precision Assessment of Tumor Changes in Response to Treatment. 
} \\
\vspace{1cm}
{\Large
Paul Tar,  Neil~Thacker, J.P.B. O'Connor et. al.\\        
\vspace{1cm}
Last updated \\
16 / 5 / 2018}                          

\vspace{9.5cm}

\includegraphics[scale=0.09]{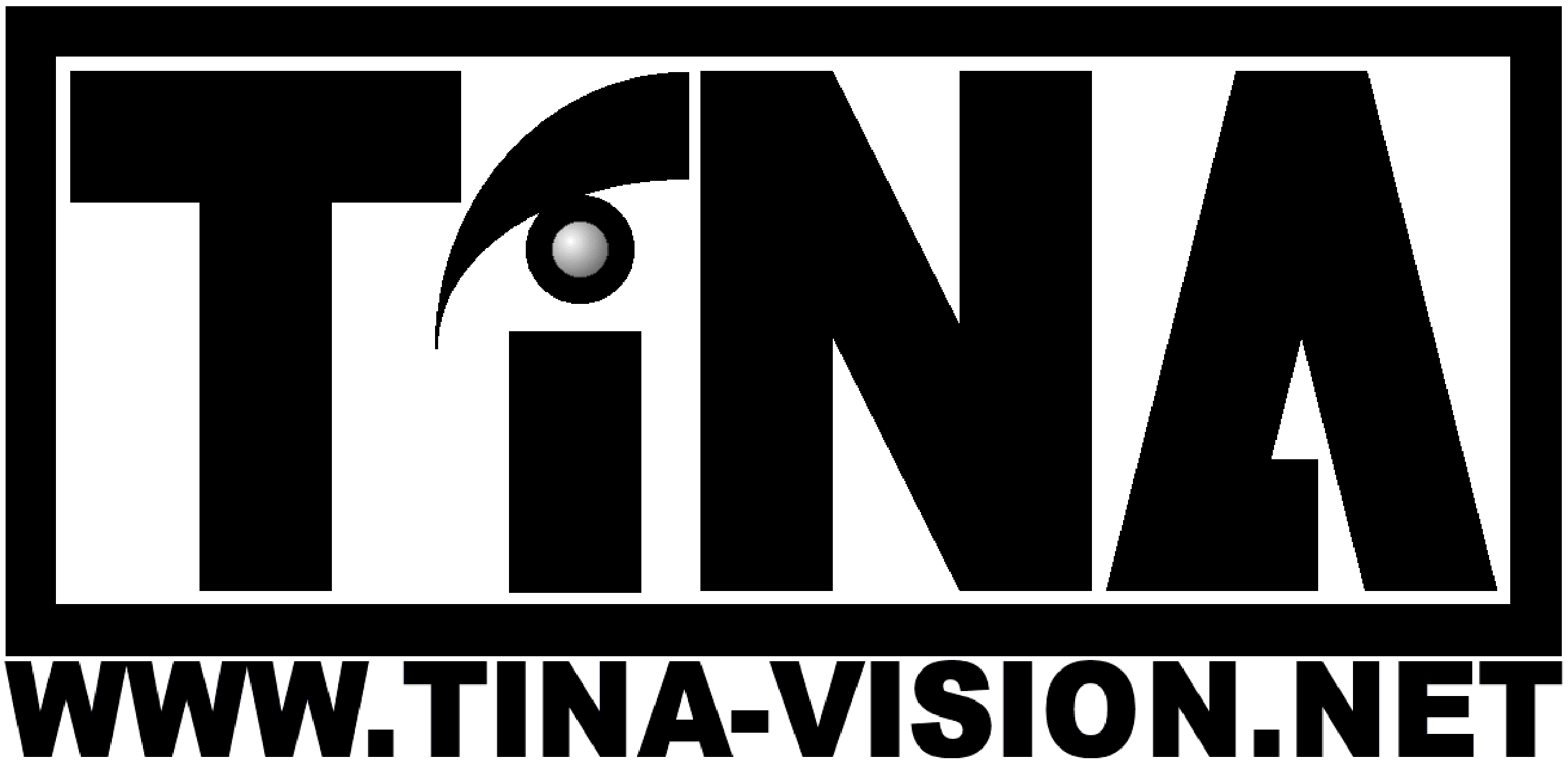}
\vspace{0.5cm}

{\large
Imaging Science and Biomedical Engineering Division, \\
Medical School, University of Manchester, \\
Stopford Building, Oxford Road, \\
Manchester, M13 9PT. \\}

\end{center}

\newpage


%


\section*{abstract}
{\textbf{Motivation:} Imaging demonstrates that preclinical and human tumors are heterogeneous, i.e. a single tumor can exhibit multiple regions that behave differently during both normal development and also in response to treatment. The large variations observed in control group tumors can obscure detection of significant therapeutic effects due to the ambiguity in attributing causes of change. This can hinder development of effective therapies due to limitations in experimental design, rather than due to therapeutic failure. An improved method to model biological variation and heterogeneity in imaging signals is described. Specifically, Linear Poisson modelling (LPM) evaluates changes in apparent diffusion co-efficient (ADC) before and 72 hours after radiotherapy, in two xenograft models of colorectal cancer. The statistical significance of measured changes are compared to those attainable using a conventional t-test analysis on basic ADC distribution parameters.
\\
\textbf{Results:} When LPMs were applied to treated tumors, the LPMs detected highly significant changes. The analyses were significant \textit{for all tumors}, equating to a gain in power of 4 fold (i.e. equivelent to having a sample size 16 times larger), compared with the conventional approach. In contrast, highly significant changes are only detected at a cohort level using t-tests, restricting their potential use within personalised medicine and increasing the number of animals required during testing. Furthermore, LPM enabled the relative volumes of responding and non-responding tissue to be estimated for each xenograft model. Leave-one-out analysis of the treated xenografts provided quality control and identified potential outliers, raising confidence in LPM data at clinically relevant sample sizes.
\\
\textbf{Availability:} TINA Vision open source software is available from www.tina-vision.net
\\
}

\section{Introduction}

\label{sec:introduction} 

Preclinical experiments and early clinical studies are essential for understanding the fundamental mechanisms driving the growth of malignant tumors and for assessing potential anti-cancer effects of new therapies (\cite{GibbsJB, ConwayJR, ClohessyJG}). In general, assessments are made by measuring tumor growth curves; by evaluating cell or plasma based assays or tissue pathology at one or more time points; and by non-invasive serial assessment by imaging. In all of these approaches, significance testing is performed typically on small numbers of subjects (\cite{ClohessyJG, WorkmanP}). This can result in low statistical power. This is especially true in cases where data is complex and variable. The limitations of small sample sizes motivate the need for efficient use of data.

Differences between groups (i.e. control versus treatment, or one thearapy versus another), are most commonly assessed using: t-tests; analysis of variance (ANOVA); or correlation analyses, which have long been available within statistical packages (\cite{KibbyMR}). These statistical approaches are used as standard within clinical and preclinical work because they facilitate estimation of confidence intervals, Z-scores and P-values. These are essential outputs for the assessment of treatment responses. However, these methods assume Gaussian distributed data, which can be impossible to corroborate using the small sample sizes often used within studies. More sophisticated modern pattern recognition approaches are experimentally applied to biomedical data for the purposes of prediction (e.g. \cite{XiaJ}), image segmentation (e.g. \cite{ZengT}) and data mining (e.g. \cite{NZong}), for example. However, they do not generally provide conventional confidence assessments (e.g. P-values etc.) and are therefore restricted to preliminary proof-of-concept rather than clinical use. 

Tumors are biologically heterogeneous, leading to numerous modes of data variability which complicate analysis (\cite{HeppnerGH, BedardPL}). Research studies using genomics (\cite{AlizadehAA}), tissue pathology (\cite{GurcanMN}) or clinical imaging (\cite{OConnorJP}) identify and quantify spatial heterogeneity and have shown that heterogeneity metrics might provide prognostic and predictive biomarkers of clinical outcome. Typically, studies measure the degree of heterogeneity within individual tumors or identify regions with certain cell populations that may mediate response to therapy and resistance (\cite{GerlingerM}). However, tumor heterogeneity can also be a practical problem for studying cancer biology. In small preclinical and clinical studies, substantial spatial variation can occur in control and treatment group tumors. This variation can obscure detection of significant biological effects of therapy, such that therapies with potential clinical benefit may be inadvertently halted in the developmental pipeline. To mitigate against this, information must be accumulated over larger sample sizes to boost statistical power, or unwanted sources of variability must be modelled.

Imaging studies generally adopt one of two approaches: One approach attempts to identify the geographic sub-regions that drive response to therapy, subsequent resistance and relapse during treatment failure. This requires solutions to the significant challenges of both image segmentation (to identify voxels with common structural or biological features) and voxel-to-voxel registration between time points. Pattern recognition techniques are often applied to solve these problems. In the presence of heterogenous control variability, it could be argued that an adverserial deep learning approach could be applied to classify treatment-affected voxels by learning the invariant characteristics of treatment, whilst ignoring the confounding changes of normal development. However, a segmented image derived through such an approach is not necessarily the best starting point for determining overall treated volumes and P-values for the significance of changes. Additionally, training a deep learning system typically requires far more data than is available in our problem domain.

In a second approach, imaging data can be regarded as a sample from a distribution, providing histograms where the spatial structure of a tumor is disregarded (\cite{JustN}). However, the complexities (e.g. non-Gaussian nature) of imaging data make it difficult to use simple histogram parameters to quantify therapy-induced changes in tumor biology (\cite{OConnorJP2}). In this approach, basic distribution parameters, such as normalisation (e.g. volume of tumor), mean values or the location of percentiles are often used in conjunction with t-tests, ANOVA etc.. This results in a large amount of information being discarded regarding the exact shape and behaviour of individual histograms.

Given the properties of our target data (histograms of Poisson samples), we sought to use linear Poisson modelling (hereafter, LPM) (\cite{TarPD, TarPD2, TarPD3}) to quantify biological variation and to model uncertainties associated with data samples acquired in clinically relevant imaging methods. LPMs can be considered as an extention to Gaussian Mixture Modelling, \cite{gmm}, where the Gaussian sub-distributions are replaced with arbitrary non-parametric probability functions. The Probability Mass Functions (PMFs) required to approximate the data are determined using an Independent Component Analysis (ICA \cite{ica}), designed for Poisson samples. LPM is a pattern recognition method specifically for quantitative work, facilitating the estimation of confidence, Z-scores and P-values.

We hypothesized that LPM would provide a method for assessing the volume of change within individual tumors, yielding a more efficient and sensitive method of detecting response to therapy, compared to conventional cohort-based analysis of imaging data. This benefit was anticipated since LPM can not only model volumetric changes - allowing estimates of the proportion of tumor changing after therapy - but can also model the effect of unwanted biological variation due to tumor growth and heterogeneity found in control data. We hypothesized that this benefit would transform the potential for image-based analyses to assess the preclinical development of novel therapeutics.

\section{Methods}
\label{sec:methodology}

Imaging data were aquired for two murine xenograft models of human colorectal cancer (LoVo and HCT116) treated with either a single high dose fraction of radiotherapy (RT) or sham (control). 8 and 13 controls were used for LoVo and HCT116, respectively. A futher 10 LoVo and 15 HCT116 treated tumors were imaged. These sample sizes are typical of those found within small preclinical trials. The MRI biomarker Apparent Diffusion Co-efficient (ADC) (\cite{PadhaniAR}) was derived for images at baseline and 72 hours after RT or sham. The tumor regions within each image were manually segmented by a clinical expert (coauthor JPB O'Connor), and the distribution of ADC values within each tumor were sampled into 2D histograms, with one axis being ADC and the other being time (t=0 and t=72), thus recording the ADC distributions pre- and post-treatment. Fig. \ref{fig:adc_images} shows example spatial distributions of such tumor data before histogramming.

{As a benchmark for comparison to our method, a conventional analysis was performed. Tumors were paired between baseline and t=72 hours, with changes in volume, changes in mean ADC value and changes in interquartile ranges computed. The per-tumor changes measured within the control groups were compared to those measured within the treatment groups using T-tests.} Additionally, LPM was used to construct a linear model of variability in the control cohorts, which were then extended to include additional variability found within the treatment cohorts. The fully-trained models were fitted to both the control and treatment groups to estimate the relative volumes associated with normal untreated tumor development and volumes associated with treatment effects. Per-tumor significances were computed, as well as cohort level significances, for comparison to the conventional t-test analyses.

Studies were performed in compliance with the NCRI Guidelines for the welfare and use of animals in cancer research (\cite{WorkmanP2}) and with Licences issued under the UK Animals (Scientific Procedures) Act 1986 (PPL 40/3212) following local Ethical Committee review.

\subsection{Tumor implantation and monitoring}

LoVo and HCT116 colorectal carcinoma cells were cultured in RPMI 1640 medium supplemented with 10\% heat inactivated fetal calf serum (FCS) at 37$^o$C in a humidified 5\% CO2 incubator. Cells were passaged every 2-3 days using TEG solution (0.25\% trypsin, 0.1\% EDTA and 0.05\% Hanks' balanced salt solution in PBS). Tumor xenografts were initiated from 5 x 106 cells per mouse (in 0.1mL serum-free culture medium) injected subcutaneous in female nu/nu CBA mice aged 10 weeks old.

Tumor size was monitored using callipers and the formula for ellipsoid volume, $V = (\pi/6)LWD$, where $L$, $W$ and $D$ are the largest orthogonal dimensions of the ellipsoid. When tumors reached 300-400 mm$^3$ in size, mice were randomised to sham or given tumor-localised RT (single 10Gy fraction) using a metal-ceramic MXR-320/36 X-ray machine (320kV, Comet AG, Switzerland). The RT was administered under ambient conditions to restrained, non-anaesthetised mice. The restrained mice were held in a lead-shielded support perpendicular to the source. Irradiation was delivered at a dose rate of 0.75 Gy/min. Mice were turned around halfway through the procedure to ensure a uniform tumor dose.  Imaging was performed at baseline immediately prior to RT and 72 hours post RT along with calliper measurement of tumor volume. After the second MRI scan, animals were killed humanely by cervical dislocation, without recovery from anaesthesia. 

\subsection{MRI acquisition and analysis}

Mice were anaesthetised with isoflurane delivered through a nose cone apparatus at 2ml/min, in 100\% oxygen gas as a carrier. Respiration rate was monitored throughout the experiment by use of an electronic respiratory monitor apparatus. A heated water bed was provided to maintain the animals at constant temperature of 36$^o$C throughout each scan. MRI was performed on a 7T Magnex instrument (Magnex Scientific Ltd, Oxfordshire, UK) interfaced to a Bruker Avance III console and gradient system (Bruker Corporation, Ettlingen, Germany), using a volume transceiver coil. Whole scan time was approximately 25 minutes per animal.

Diffusion-weighted imaging (TR/TE = 2250/20ms; $\alpha = 90^o$; b values 150, 500 and 1000 s/mm$^2$ along one diffusion direction; matrix 128 x 128 and FOV 2.56 x 2.56cm; 15 contiguous slices of 0.6mm thickness) was performed after localisation with a T2-weighted anatomical sequence (TR/TE = 2410/50; $\alpha = 136.8^o$; matrix 256 x 256 and FOV 2.56 x 2.56cm; 15 contiguous slices of 0.6mm thickness). ADC maps were generated by selecting a region of interest on the lowest b value image. Voxel-wise values of ADC (Supplementary Data File) were calculated using in house software across the tumor using a least squares fitting routine for the equation $S = S0e^{-bD}$, where $S­0$ represents the signal intensity in the absence of a diffusion sensitising gradient, $S$ the signal intensity for a particular $b$ value, $b$ the numerical value in $s$/mm$^2$ and $D$ the apparent diffusion coefficient (mm$^2$/s). 

To validate the ADC measurement in this protocol, measurements were verified using an ice water phantom, consisting of an inner chamber of ice water surrounded by a larger chamber of ice to maintain the inner chamber water at approximately $0^o$C (\cite{DoblasS}).

Baseline and change in tumor volume and ADC (mean value and IQR) parameters were compared between control and treated tumors using Student's t test for independent samples in IBM SPSS Statistics v.22 (Armonk, NY). All tests were two tailed. These tests were performed and combined to provide comparison with the statistics derived from LPM (see below). In all tests, p<0.05 was considered to indicate statistical significance. Corrections for multiple comparisons were applied where necessary. 

\subsection{Linear Poisson modelling of ADC data}

A linear Poisson model describes a set of histograms (i.e. ADC distributions) using a linear combination of PMFs, where each PMF represents some sub-component {(e.g. a mode of variability/behaviour)} of the signal:

\begin{equation}
\label{equ:LPM}
\mathbf{H}(ADC,t) \approx \sum_C^{N_C} P(ADC,t|C)\mathbf{Q}_{C} + \sum_T^{N_T} P(ADC,t|T)\mathbf{Q}_{T}
\end{equation}

where $\mathbf{H}(ADC,t)$ is the histogram bin recording the frequency of observed ADC values within range $ADC$ at time $t$; {$C$ is a label indicating a component of control behaviour; $T$ is a label indicating a component of treatment behaviour, as determined by the \textit{additional variability} within the treatment group, i.e. the behaviour in treated cases that cannot be accounted for already by control behaviour}; $P(ADC,t|C)$ and $P(ADC,t|T)$ are the probabilities of observing an ADC value in range at time $t$ from control behaviour or treatment behaviour; and $\mathbf{Q}_{C}$ and $\mathbf{Q}_{T}$ are the quantities of each component in the data. There are $N_C$ control components and $N_T$ treatment components. Each component can broadly be considered as a type of tissue development in control or treatment, corresponding to a mode of heterogeneous variability. The more complex a tumor and its response to treatment, the greater the number of components the tumor needs in its model.

{Given a set of control tumors, $i \in \{1, 2, \ldots, S_C\}$ (where $S_C$ is the control cohort sample size), and treated tumors, $j \in \{1, 2, \ldots, S_T\}$ (where $S_T$ is the treatment cohort sample size), an LPM is used to provide Likelihood solutions to PMFs and quantities. Estimation of quantities and probabilities are achieved using Expectation Maximisation to optimise the following Extended Maximum Likelihood for control cohorts

\begin{equation}
\label{equ:likelihood_a}
\ln \mathcal{L} = \sum_{i,ADC,t} \ln \left[\sum_C^{N_C} P(ADC,t|C)\mathbf{Q}_{Ci}\right] \mathbf{H}_i(ADC,t) - \sum_C \mathbf{Q}_{Ci} 
\end{equation}

\noindent and the following for treatment cohorts

\begin{equation}
\label{equ:likelihood_b}
\ln \mathcal{L} = \sum_{j,ADC,t} \ln \left[\sum_C^{N_C} P(ADC,t|C)\mathbf{Q}_{Cj} + \sum_T^{N_T} P(ADC,t|T)\mathbf{Q}_{Tj} \right] 
\end{equation}

\[
\times \mathbf{H}_j(ADC,t) - \sum_C \mathbf{Q}_{Cj} - \sum_T \mathbf{Q}_{Tj}
\]

Thus, the model is trained in two parts. Initially, $N_C$ terms are estimated using only the control cohort as training data (Equ. \ref{equ:likelihood_a}). Once the PMFs for control behaviour have been learnt, these components are automatically included as modes of behaviour within the treatment cohort. The additional $N_T$ components that describe the extra variability expected due to treatment are then learnt using the treatment cohort, keeping the original $N_C$ components as part of the model (Equ. \ref{equ:likelihood_b}). In this way, parts of a treated tumor's ADC distribution can be partitioned into quantities of responding and non-responding behaviour.}

{A model selection process identifies the optimum number of LPM components required to describe the ADC distributions. Multiple models are constructed with increasing numbers of components, with the best fitting models being selected for use in subsequent analysis. The number of components required to describe each class of response, i.e. $N_C$ for control and $N_C+N_T$ for treatment, is determined by adding additional components until the $\chi^2$ per degree of freedom between LPM and ADC histograms reaches a minimum, ideally at unity:}

\begin{equation}
\label{equ:chi2D}
\chi^2_D = \frac{1}{D} \sum_{ADC,t} \frac{ [\sqrt{\mathbf{H}(ADC,t)} - \sqrt{\mathbf{M}(ADC,t)}]^2 }{\sigma^2_{ADC,t}}
\end{equation}

where $D$ is the number of degrees of freedom and $\sigma^2$ is the variance predicted on the residual. The square-roots are present to transform the Poisson distributed histogram frequencies into Gaussian-like variables to improve this figure of merit's approximation to ideal $\chi^2$ statistics, as described in \cite{Anscombe}.

Assuming independent Poisson errors ($\sigma^2_{\mathbf{H}} \approx \mathbf{H}$), LPMs provide estimates of uncertainties by summing the effects of individual Poisson bins into quantity error covariances. This is achieved using error propagation. The error covariance can be further scaled by $\chi^2_D$ (goodness-of-fit) computed from LPM-data residuals to boost errors to better match actual distributions of true residuals, i.e. scaling factor that can be caused by the up-sampling of MRI data. A covariance matrix for quantities, $\mathbf{Q}$, can be estimated using:

\begin{equation}
\label{equ:cov_stat}
\mathbf{C}_{ij} = \chi^2_D \sum_m \left [ \left (\frac{\partial \mathbf{Q}_i}{\partial \mathbf{H}(ADC,t)} \right ) \left ( \frac{\partial \mathbf{Q}_j}{\partial \mathbf{H}(ADC,t)} \right ) \sigma^2_{\mathbf{H}} \right] 
\end{equation} 

where $\mathbf{C}$ is the error covariance matrix for the estimated quantities.  

The statistical significance of treatment response is computed by dividing the sum of treatment quantities $\sum_T \mathbf{Q}_T$ by the estimated error on that total quantity. This provides a Z-score, indicating how many standard deviations from zero the response is estimated to be.

\subsection{Model validation}

The null hypothesis from which P-values are computed is that behaviour is consistent with control, and that control behaviour is predictable. This behaviour must generalises to unseen control data. In contrast, treatment behaviour only needs to be different from control. The validity of this null hypothesis relies upon there being no significant changes in independent non-treatment groups. We used a combination of control and leave-one-out testing to provide technical validation.

Treatment models were fitted to control training data to ensure the measured effects of treatment were consistent with zero (with error bars). Additionally, if control LPM are representative of typical non-treated tumors, then their application to independent data should yield equivalent results to data from which the models were original estimated. A leave-one-out analysis was therefore performed in which multiple models were constructed, with each control tumor being excluded in turn, before being assessed as an independent sample. This leave-one-out strategy in control data enables stringent testing to be performed in numbers of data sets that are typical of those used in preclinical cancer imaging experiments (\cite{BernsenMR}). This approach also protects against false-positive results through quality control (i.e. representativeness testing) of training data.

\section{Results}
\label{sec:results}

\begin{figure}[!tpb]
\centerline{\includegraphics[width=0.8\linewidth]{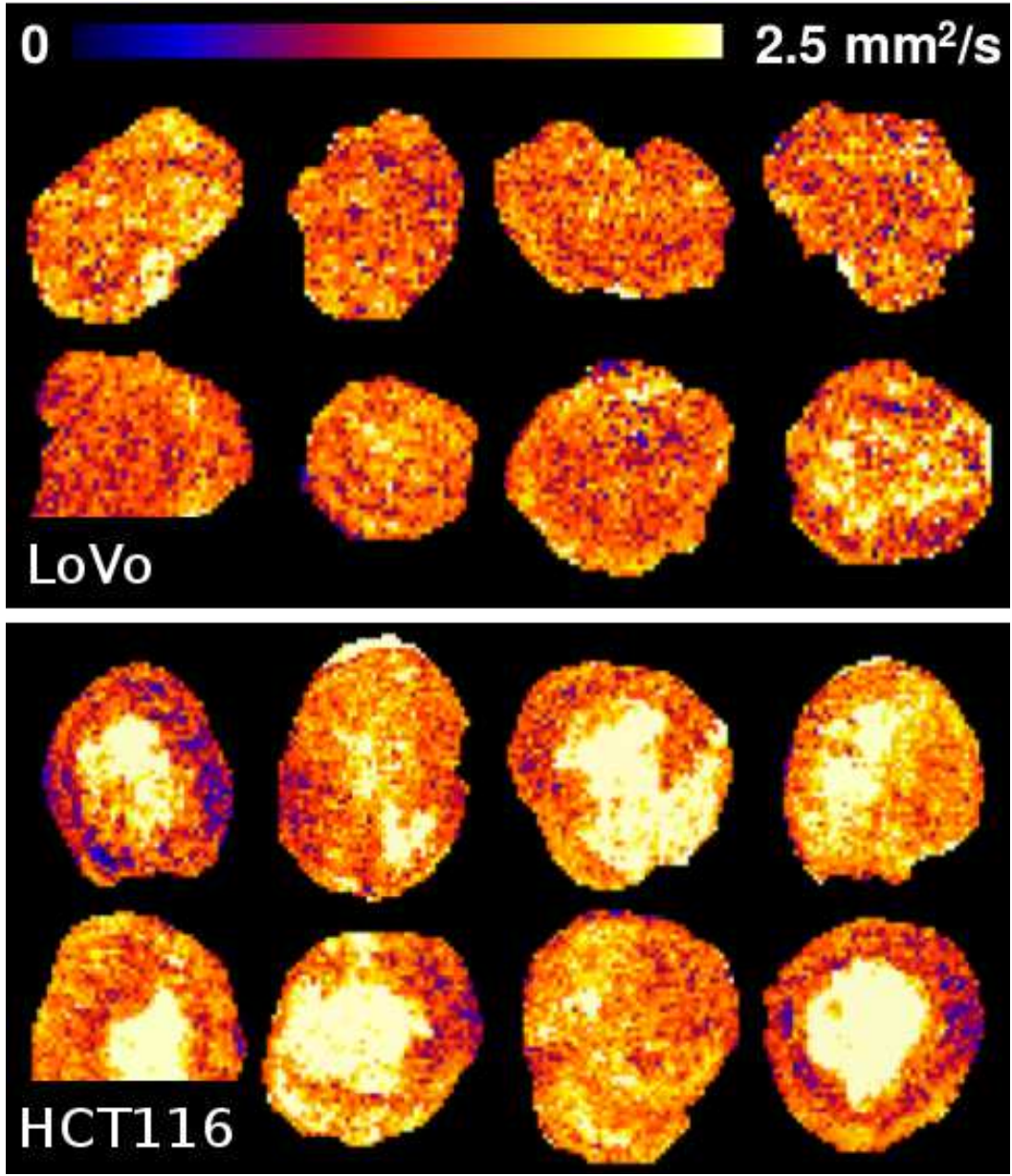}}
\caption{Example spatial distributions of ADC values in selected tumors. Visually, HCT116 tumors are more complex and variable than LoVo tumors.}\label{fig:adc_images}
\end{figure}

\begin{figure}[!tpb]
\centerline{\includegraphics[width=1.0\linewidth]{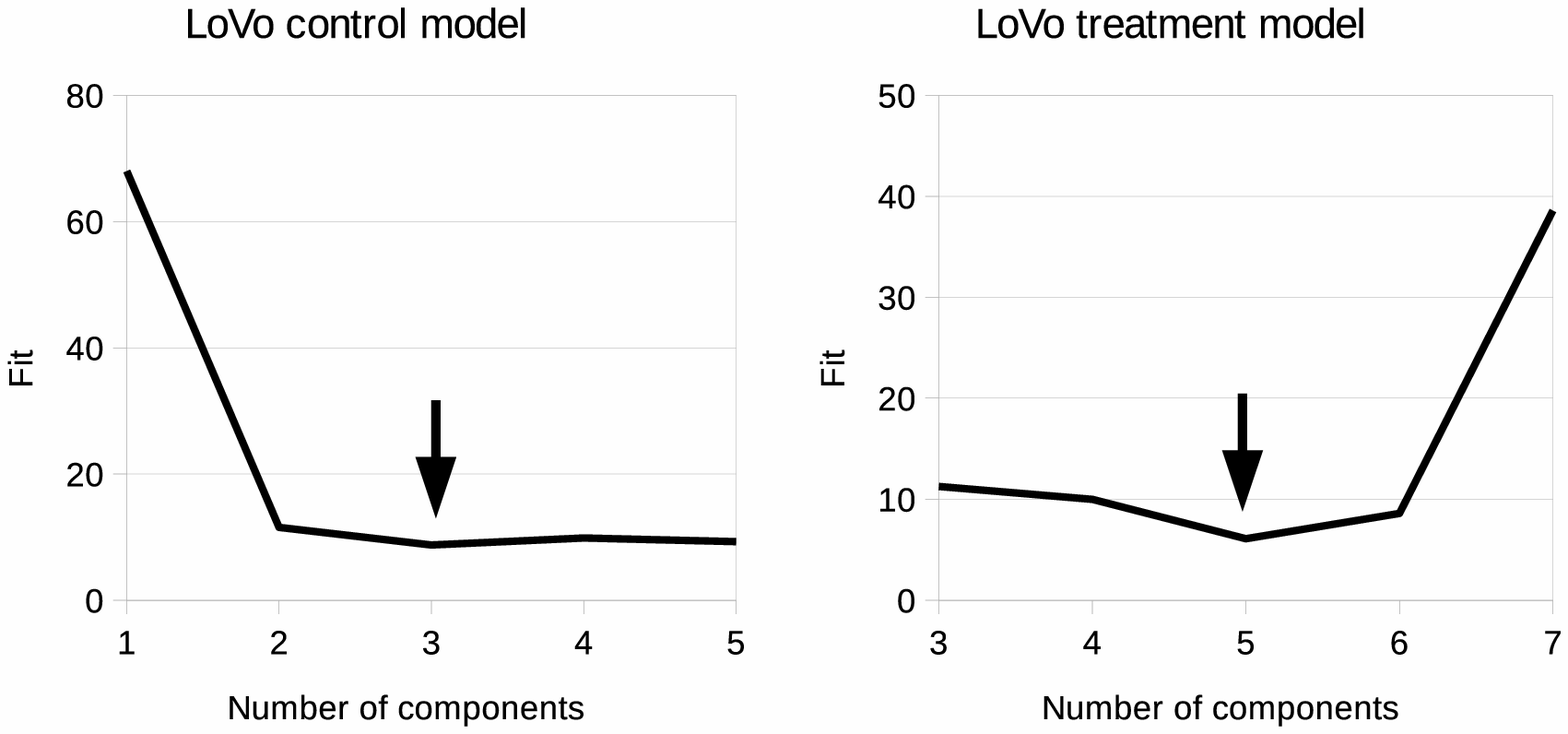}}
\caption{Model selection curves indicating necessary number of components to describe control and treatment groups. Left: $\chi^2_D$ as a function of $N_C$ for LoVo. Right: $\chi^2_D$ as a function of $N_C+N_T$ for LoVo.}\label{fig:lovo_model_selection}
\end{figure}

\begin{figure}[!tpb]
\centerline{\includegraphics[width=1.0\linewidth]{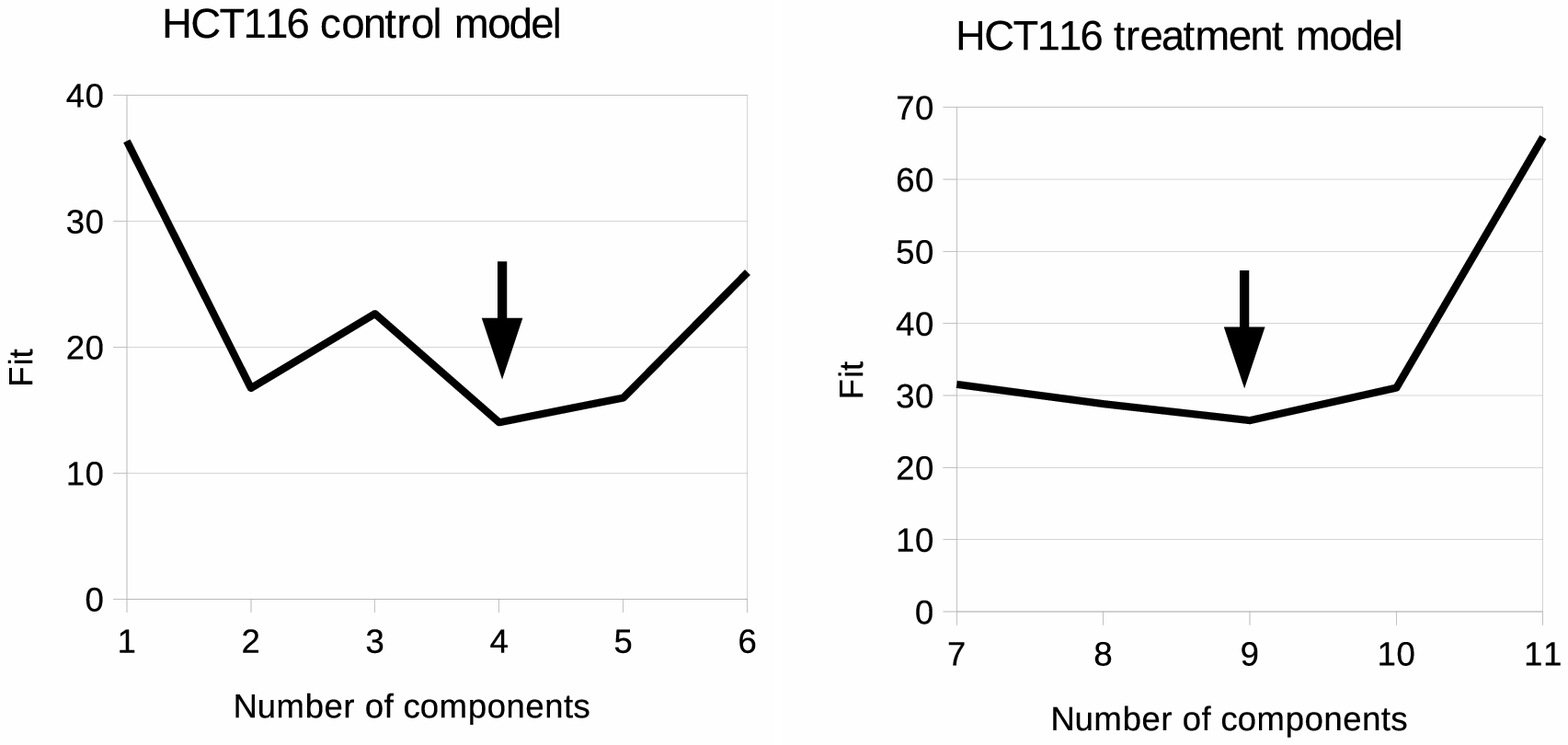}}
\caption{Model selection curves indicating necessary number of components to describe control and treatment groups. Left: $\chi^2_D$ as a function of $N_C$ for HCT116. Right: $\chi^2_D$ as a function of $N_C+N_T$ for HCT116.}\label{fig:hct_model_selection}
\end{figure}

\begin{figure}[!tpb]
\centerline{\includegraphics[width=0.9\linewidth]{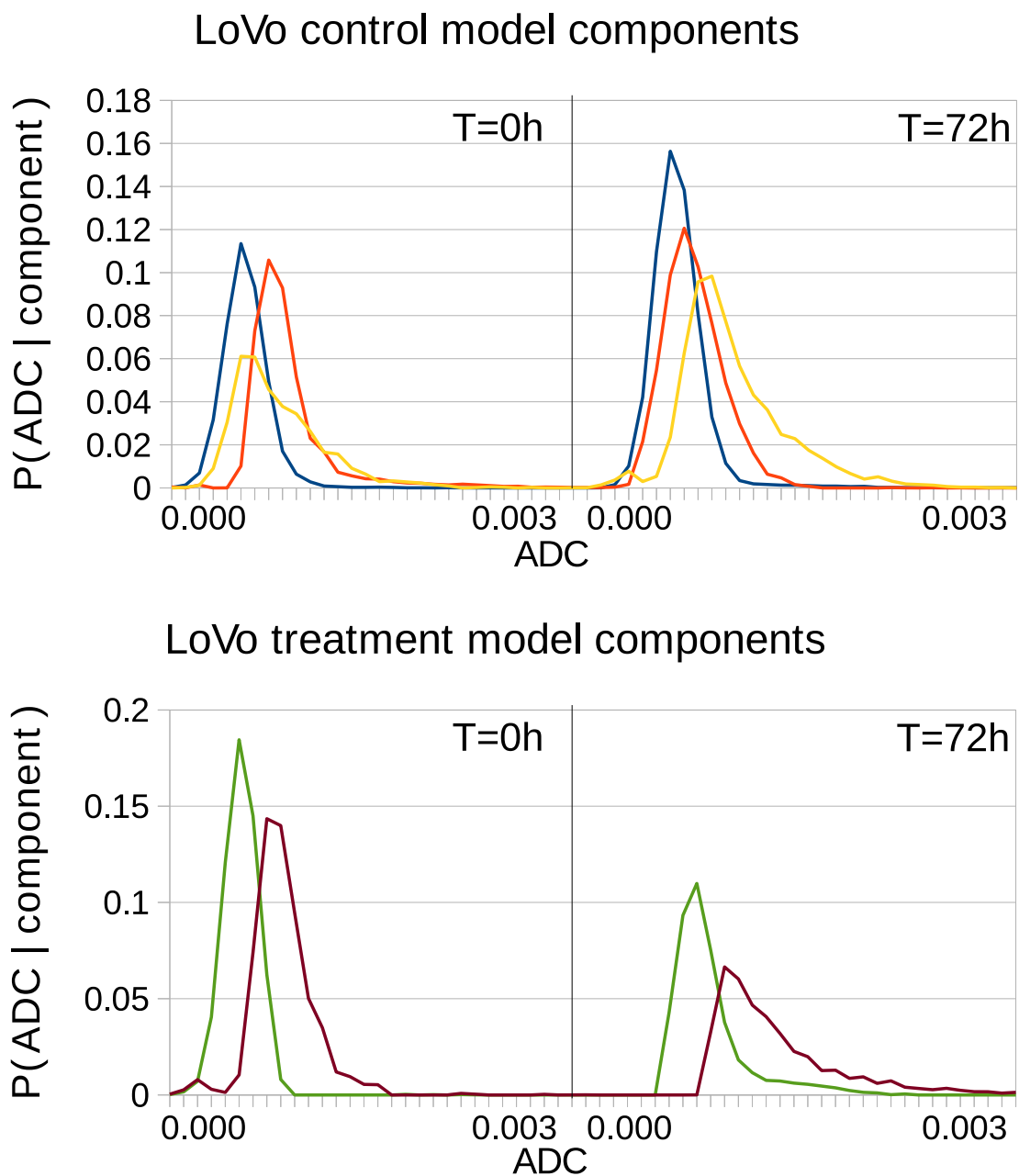}}
\caption{Estimated components (PMFS: $P(ADC,t|C)$ and $P(ADC,t|T)$), one color per component. Left and right plots indicate baseline and 72 hours. Top: LoVo control components. Bottom: LoVo treatment components}\label{fig:lovo_components}
\end{figure}

\begin{figure}[!tpb]
\centerline{\includegraphics[width=0.9\linewidth]{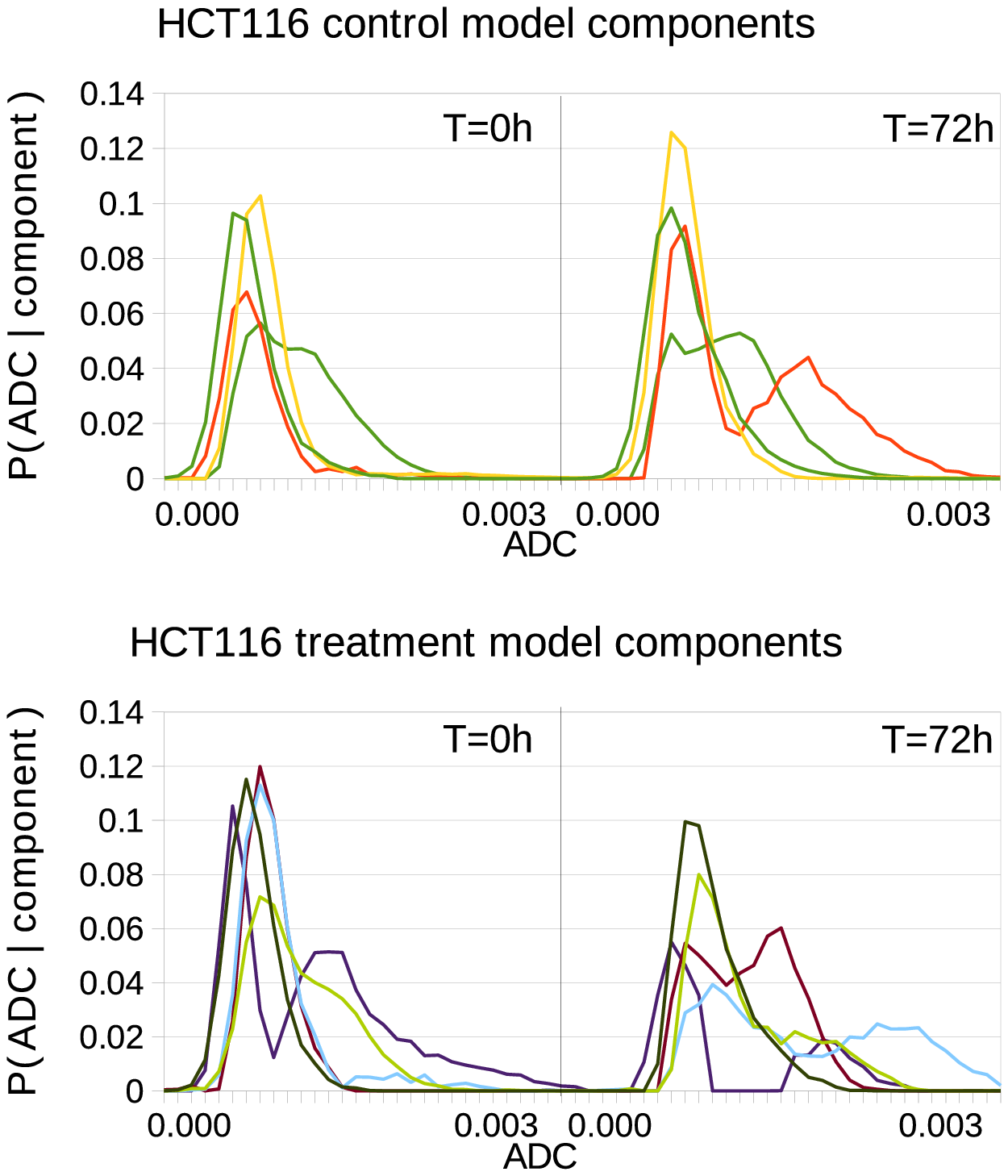}}
\caption{Estimated components (PMFS: $P(ADC,t|C)$ and $P(ADC,t|T)$), one color per component. Left and right plots indicate baseline and 72 hours. Top: HCT116 control components. Bottom: HCT116 treatment components}\label{fig:hct_components}
\end{figure}

\begin{figure}[!tpb]
\centerline{\includegraphics[width=1.0\linewidth]{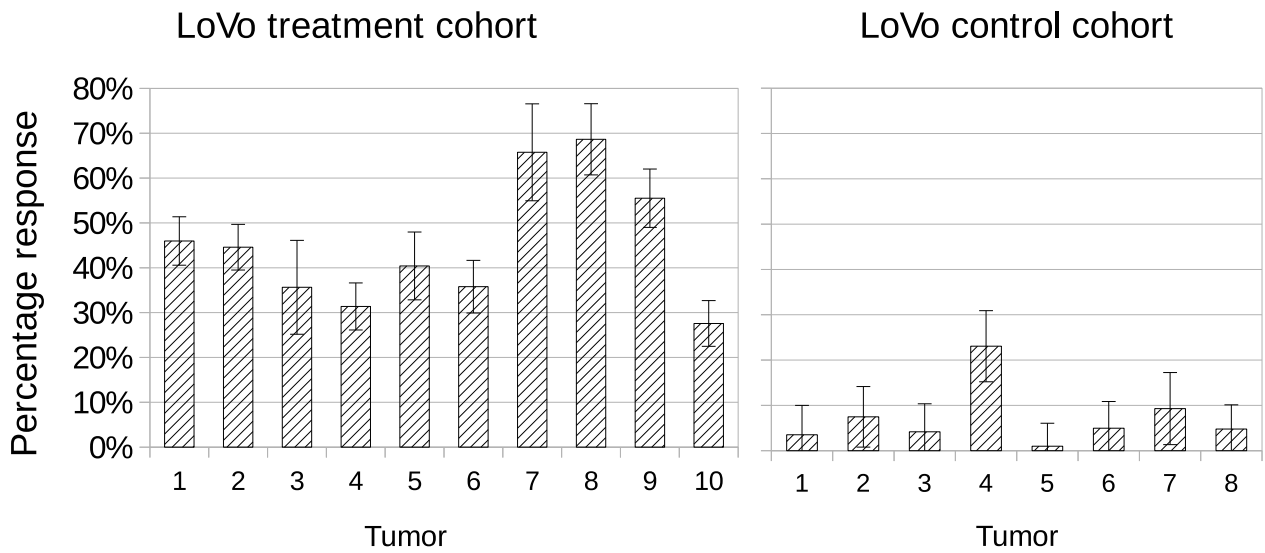}}
\caption{Volume response to treatment (i.e. $\sum_T \mathbf{Q}_T$) for LoVo tumors. Left: Treatment cohort, with significant non-zero values. Right: Control cohort, with values consistent with zero (i.e. within level of predicted error) with possible outlier at tumor 4. All error bars show $\pm$ 1 standard deviation.}\label{fig:lovo_response}
\end{figure}

\begin{figure}[!tpb]
\centerline{\includegraphics[width=1.0\linewidth]{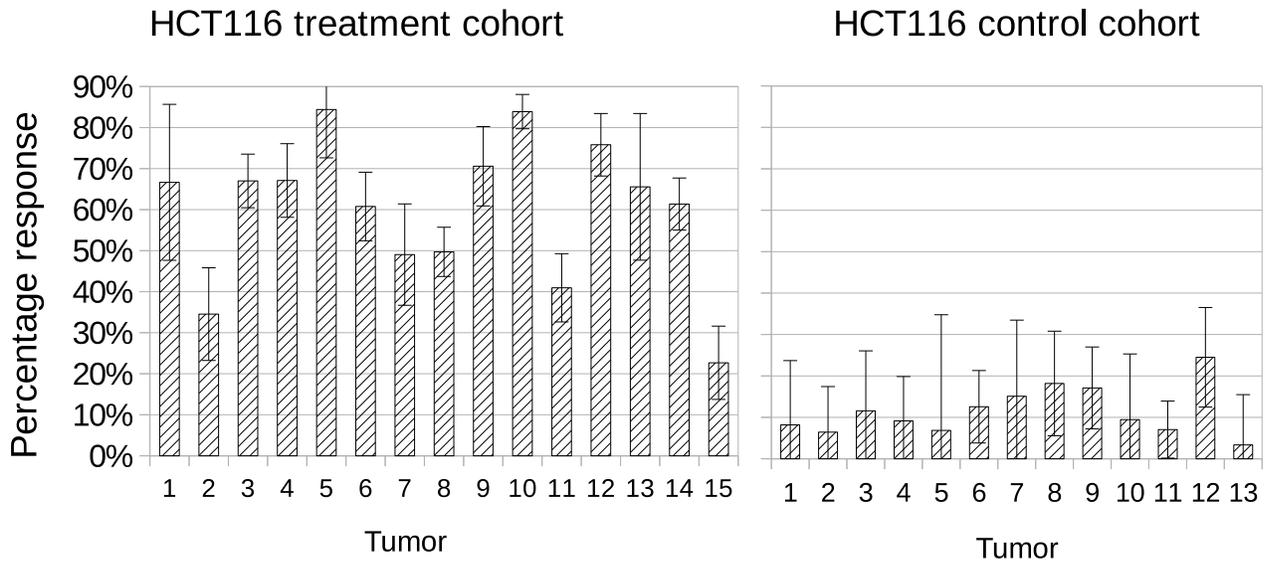}}
\caption{Volume response to treatment (i.e. $\sum_T \mathbf{Q}_T$) for HCT116 tumors. Left: Treatment cohort, with significant non-zero values. Right: Control cohort, with values consistent with zero (i.e. within level of predicted error). All error bars show $\pm$ 1 standard deviation.}\label{fig:hct_response}
\end{figure}

\begin{table}[!tpb]
\centering
\caption{Conventional t-test analysis. Note that the cohort level significances are 5.2 standard deviations change and 8.1 standard deviations change for LoVo and HCT116, respectively. These figures should be compared to the cohort level significances of Tables \ref{tab:lovo_treatment_sigs} and \ref{tab:hct_treatment_sigs}}
\label{tab:lovo_t_tests}
\begin{tabular}{lcrrr}

\hline
\textbf{Measurement} & \textbf{Z score} & \textbf{P value} & \\
\hline
LoVo vol. change & 3.3 & 0.001 \\
LoVo mean ADC change & 3.5 & 0.0004 \\
LoVo IQR change & 2.0 & 0.041 \\
\textbf{LoVo combined} & \textbf{5.2} & \textbf{ $\le$ 0.000001} \\
\hline
HCT116 vol. change & 4.6 & 0.0008 \\
HCT116 mean ADC change & 4.3 & 0.0009 \\
HCT116 IQR change & 2.1 & 0.047 \\
\textbf{HCT116 combined} & \textbf{8.1} & \textbf{$\le$ 0.000001} \\
\hline

\end{tabular}
\end{table}

\begin{table}[!tpb]
\centering
\caption{LoVo treatment cohort result significances. Note that the cohort-level significance (bottom row) is approximately 4 times that for LoVo in Table \ref{tab:lovo_t_tests}}.
\label{tab:lovo_treatment_sigs}
\begin{tabular}{ccrrr}

\hline
\textbf{Tumor} & \textbf{Z score} & \textbf{P value} & \% \textbf{Effect} & \% \textbf{Error} \\
\hline
1 & 8.5 & $\le$ 0.000001 & 45.98\% & 5.40\% \\
2 & 8.8 & $\le$ 0.000001 & 44.59\% & 5.09\% \\
3 & 3.4 &  0.000667 & 35.64\% & 10.47\% \\
4 & 5.9 & $\le$ 0.000001 & 31.37\% & 5.24\% \\
5 & 5.3 & $\le$ 0.000001 & 40.41\% & 7.57\% \\
6 & 6.1 & $\le$ 0.000001 & 35.77\% & 5.87\% \\
7 & 6.1 & $\le$ 0.000001 & 65.75\% & 10.80\% \\
8 & 8.6 & $\le$ 0.000001 & 68.64\% & 7.94\% \\
9 & 8.5 & $\le$ 0.000001 & 55.51\% & 6.50\% \\
10 & 5.4 & $\le$ 0.000001 & 27.57\% & 5.10\% \\
\textbf{Combined} & \textbf{21.8} & \textbf{ $\le$ 0.000001}\\
\hline

\end{tabular}
\end{table}

\begin{table}[!tpb]
\centering
\caption{HCT116 treatment cohort result significances. Note that the cohort-level significance (bottom row) is approximately 4 times that for HCT116 in Table \ref{tab:lovo_t_tests}}
\label{tab:hct_treatment_sigs}
\begin{tabular}{ccrrr}

\hline
\textbf{Tumor} & \textbf{Z score} & \textbf{P value} & \% \textbf{Effect} & \% \textbf{Error} \\
\hline
1 & 3.5 & 0.000453 & 66.66\% & 19.01\% \\
2 & 3.1 & 0.002239 & 34.53\% & 11.30\% \\
3 & 10.3 & $\le$ 0.000001 & 67.00\% & 6.53\% \\
4 & 7.5 & $\le$ 0.000001 & 67.10\% & 8.97\% \\
5 & 7.2 & $\le$ 0.000001 & 84.41\% & 11.80\% \\
6 & 7.3 & $\le$ 0.000001 & 60.77\% & 8.35\% \\
7 & 3.9 & 0.000072 & 49.04\% & 12.36\% \\
8 & 8.3 & $\le$ 0.000001 & 49.71\% & 6.01\% \\
9 & 7.3 & $\le$ 0.000001 & 70.58\% & 9.67\% \\
10 & 20.1 & $\le$ 0.000001 & 83.89\% & 4.18\% \\
11 & 4.9 & $\le$ 0.000001 & 40.94\% & 8.32\% \\
12 & 9.9 & $\le$ 0.000001 & 75.82\% & 7.61\% \\
13 & 3.7 & 0.000243 & 65.56\% & 17.87\% \\
14 & 9.7 & $\le$ 0.000001 & 61.36\% & 6.35\% \\
15 & 2.5 & 0.0111474 & 22.67\% & 8.93\% \\
\textbf{Combined} & \textbf{32.6} & \textbf{$\le$ 0.000001}\\
\hline

\end{tabular}
\end{table}

\begin{table}[!tpb]
\centering
\caption{LoVo control cohort result significances. Main figures show results for leave-all-in analysis. Figures in brackets show leave-one-out results, where the model was trained on all except the current tumor before being applied to the current tumor.}
\label{tab:lovo_control_sigs}
\begin{tabular}{cccrrrrrr}

\hline
\textbf{Tumor} & \textbf{Z} & \textbf{(Z)} & \textbf{P} & \textbf{(P)} & \% \textbf{Effect} & \textbf(\%) & \% \textbf{Error} & \textbf(\%) \\
\hline
1 & 0.5 & (1.2) & 0.58 & (0.20) & 3.51\% & (23.92) & 6.48\% & (18.86) \\
2 & 1.1 & (1.0) & 0.26 & (0.31) & 7.47\% & (8.89) & 6.69\% & (8.75) \\
3 & 0.6 & (0.5) & 0.49 & (0.57) & 4.17\% & (5.95) & 6.18\% & (10.73) \\
4 & 2.9 & (5.2) & 0.00 & (0.00) & 23.05\% & (32.73) & 7.84\% & (6.29) \\
5 & 0.2 & (0.0) & 0.84 & (0.97) & 0.98\% & (0.38) & 5.07\% & (11.83) \\
6 & 0.8 & (0.3) & 0.40 & (0.72) & 4.95\% & (9.24) & 5.89\% & (26.15) \\
7 & 1.2 & (0.4) & 0.24 & (0.68) & 9.28\% & (6.10) & 7.96\% & (15.17) \\
8 & 0.9 & (0.5) & 0.37 & (0.56) & 4.78\% & (7.99) & 5.34\% & (13.76) \\
\hline

\end{tabular}
\end{table}

\begin{table}[!tpb]
\centering
\caption{HCT116 control cohort result significances. Main figures show results for leave-all-in analysis. Figures in brackets show leave-one-out results, where the model was trained on all except the current tumor before being applied to the current tumor.}
\label{tab:hct_control_sigs}
\begin{tabular}{cccrrrrrr}

\hline
\textbf{Tumor} & \textbf{Z} & \textbf{(Z)} & \textbf{P} & \textbf{(P)} & \% \textbf{Effect} & \textbf(\%) & \% \textbf{Error} & \textbf(\%) \\
\hline
1 & 0.5 & (0.1) & 0.59 & (0.86) & 8.18\% & (10.29) & 15.53\% & (59.55) \\
2 & 0.6 & (0.3) & 0.55 & (0.70) & 6.43\% & (7.41) & 11.00\% & (19.52) \\
3 & 0.8 & (1.0) & 0.42 & (0.29) & 11.57\% & (23.31) & 14.48\% & (22.07) \\
4 & 0.9 & (0.7) & 0.39 & (0.44) & 9.14\% & (7.97) & 10.70\% & (10.53) \\
5 & 0.2 & (0.3) & 0.80 & (0.71) & 6.84\% & (13.61) & 27.93\% & (36.93) \\
6 & 1.4 & (1.8) & 0.15 & (0.58) & 12.55\% & (29.65) & 8.74\% & (15.65) \\
7 & 0.8 & (0.5) & 0.41 & (0.56) & 15.10\% & (16.67) & 18.36\% & (29.09) \\
8 & 1.4 & (2.2) & 0.14 & (0.02) & 18.18\% & (36.09) & 12.62\% & (16.05) \\
9 & 1.7 & (1.4) & 0.08 & (0.16) & 17.09\% & (20.52) & 9.89\% & (14.62) \\
10 & 0.6 & (0.8) & 0.54 & (0.37) & 9.45\% & (13.43) & 15.84\% & (15.19) \\
11 & 1.0 & (0.8) & 0.30 & (0.39) & 7.04\% & (20.15) & 6.89\% & (23.91) \\
12 & 2.0 & (3.5) & 0.04 & (0.00) & 24.50\% & (36.00) & 12.01\% & (10.10) \\
13 & 0.3 & (0.0) & 0.78 & (0.96) & 3.34\% & (2.91) & 12.14\% & (69.60 \\
\hline

\end{tabular}
\end{table}


\subsection{Cohort volumetrics and summary ADC detect RT response}

Volume and basic ADC distribution parameters demonstrated that significant growth inhibition was induced by RT in both xenograft models at 72 hours, relative to control. In both LoVo and HCT116, RT reduced volume, increased mean ADC value and increased IQR of the ADC distribution, relative to control. Treatment effects were detected at the \textit{cohort level}, as summarised in Table \ref{tab:lovo_t_tests}, reaching high levels of significance ($p < 0.000001$). The Z-scores and P-values were computed from t-tests on the three parameters individually (changes in volume, mean and IQR). The combined Z-score and P-value values show the significances attainable when the three parameters are considered jointly, assuming each provides independence evidence of change. As t-tests are applied to the group, individual tumor change assessments are not possible using this method.

\subsection{LPM identifies the varying complexity of different xenograft models}

For each xenograft model, an LPM was constructed independently and the number of model components was selected on the basis of leave-one-out cross validation. This yielded 3 components to describe ADC distributions in the LoVo control tumors, with an additional 2 required for the variability caused by treatment. An equivalent and independent process was performed for the HCT116 tumors. This yielded 4 components in control tumors and an additional 5 for treatment response.

{The plots in Fig. \ref{fig:lovo_model_selection} and \ref{fig:hct_model_selection} show these results in detail in terms of goodness-of-fits ($\chi^2_D$) for models with different numbers of components. We seek the number of components which gives the minimum. The best solutions are indicated with arrows: LoVo $N_C = 3$ and $N_T = 2$ ($N_C+N_T=5$ in the plot); HCT116 $NC = 4$ and $N_T=5$ ($N_C+N_T = 9$ in the plot).}

The HCT116 tumors are expected to be more complex than LoVo, as they show a greater inter-quartile range of ADC values and can be seen to be more heterogeneous upon visual inspection. A more complex tumor is expected to require a greater number of LPM components to be modelled. The LPM data indicates that the HCT116 xenografts were more spatially complex than the LoVo xenografts and that LPM can detect this differing level of tumor complexity, which is expected of these particular tumors. The greater number of components required to describe HCT116 tumors reflects the higher variability that can be seen visually in Fig. \ref{fig:adc_images}. 

The ADC distributions associated with the extracted components can be seen in Fig. \ref{fig:lovo_components} and Fig. \ref{fig:hct_components}. Each component is a probability distribution, showing the statistical correlations between ADC values between the two time points. These correspond to the $P(ADC,t|T)$ and $P(ADC,t|C)$ parts of the model. The weighted sum of these distributions describe the variability observed within the data. Biologically, each component can be interpreted as a sub-population of ADC values found within the tumours. The higher ADC values at t=72 are more probable, indicating greater diffusion due to less restricted fluid movement.

\subsection{LPM validation identifies outliers in control groups}

{We used control testing and a leave-one-out approach to validate the ability of the model to distinguish data with different ADC distributions to ensure control growth was corrected accounted for. To do this, we applied fully trained models (i.e. leave-all-in) to both LoVo and HCT116 control data, followed by reduced models where each tumor in turn is excluded before being used as an independent test data point (i.e. leave-one-out). Responses were computed in each case with leave-all-in results plotted in the right of Fig. \ref{fig:lovo_response} and \ref{fig:hct_response}. Leave-all-in and leave-one-out results are directly compared in Table \ref{tab:lovo_control_sigs} and \ref{tab:hct_control_sigs}.

A Z score is given by diving the size of a response by the 1 standard deviation error on that response.} In cohorts of around 10, all control tumors would be expected to have Z scores of less than 2. This was found for all but two tumors (LoVo 4 and HCT116 12), with average Z scores from full models of 1.05 for LoVo and 0.94 for HCT116. Differences between alternative models (leave-all-in and each of the various leave-one-out possibilities) were statistically equivalent, implying that estimated volumes were the same, within limits of estimated errors. These data show that the model performs as expected, correctly accounting for each control tumor distribution as being constructed of components from untreated voxel values.

The leave-one-out approach not only validates the LPM method, but also identifies outlier data in the control cohort. {During full analysis LoVo control tumor 4 showed a Z score of 2.9 for estimated treatment volume and HCT116 control tumor 12 showed a Z score of 2.0. These increased to 5.2 and 3.5, respectively, for leave-one-out analysis, implying differences from other control data.} This could be explained by the data being an atypical, yet otherwise valid, control sample, which could have been better modelled using additional training data. For the current study, we elected to leave these data in the control group, to impose a 'worst case scenario` on our data, since we are describing a new methodology. More reasonably, this can be explained by these two control tumors being outliers.

Therefore, LPM with leave-one-out validation enables statistically robust identification of outliers in control data, which can be a critical step in avoiding equivocal results in small low-powered preclinical studies. 

\subsection{LPM quantifies the percentage responding volume in each tumor}

Non-responding tumor was defined by the sum of the control model component volumes ($\sum_C \mathbf{Q}_C$) and responding tumor was defined by the sum of the treated model component volumes ($\sum_T \mathbf{Q}_T$). The proportion of tumor changing with therapy was calculated, along with error bars (right of Fig. \ref{fig:lovo_response} and \ref{fig:hct_response}). All LoVo and HCT116 tumors treated with RT showed statistically significant volumes of responding tumor, i.e. the responding volue was above zero, beyond the level expected by noise alone. For LoVo, proportion of volume responding to RT varied between 27.6 to 68.6\% (median responding volume 40.4\%). For HCT116, proportion of volume responding to RT varied between 22.7 to 84.4\% (median responding volume 61.4\%). In comparison, all control tumors (except outlier LoVo control tumor 4) had responding volumes consistent with zero. These measurements are possible with the LPM method, but not the t-test method. 

\subsection{LPM biomarkers of response are more powerful than conventional analyses}

In LPM, the error estimates on measured affected volumes incorporate systematic processes associated with learning the model parameters (i.e. determination of PMFs), as well as the statistical errors on weighting factors used to describe each case. LPM can capture the uncertainties on the distribution components and the weighting factors using the error estimates provided by the method. This enables construction of hypothesis tests for individual data sets, by testing the null hypothesis (i.e. zero response) on a case by case basis. The probability of the treatment volume being consistent with zero on the basis of estimated error was measured. Tables \ref{tab:lovo_treatment_sigs} and \ref{tab:hct_treatment_sigs} show the individual and cohort significances. {Tables \ref{tab:lovo_control_sigs} and \ref{tab:hct_control_sigs} show control cohort responses for comparison.}

The LPM approach implicitly combines information from volume and ADC change. To ensure a fair comparison between LPM and conventional measures, we combined the significance for conventional volume and ADC (mean and IQR), giving a total Z score of 5.2 standard deviations for LoVo (Table \ref{tab:lovo_t_tests}).  LPM results showed higher Z score and more significant p values for many of the individual treated tumors compared to the conventional cohort-level statistics for imaging biomarkers. 

The combined Z score from the LPM was 21.8 standard deviations for LoVo tumors. Since a linear increase in Z score requires a quadratic increase in data quantity, approximately 17-18 times more data (square of 21.8/5.2) would be needed for LoVo tumors to demonstrate the same treatment effect with equivalent power using volume and mean ADC compared to LPM. This equates to an increase in power of approximately 4 fold. An equivalent comparison of summary statistics and LPM statistics in HCT116 xenografts treated with RT showed a similar gain in statistical power. These data reveal that mathematical modelling of imaging data through LPM enables substantial increase in statistical power to detect response to therapy.

\section{Discussion}
\label{sec:discussion}

In this study we describe how modelling the spatial heterogeneity present in imaging data can increase statistical power of identifying response to therapy. We investigated a technique called linear Poisson modelling in a well understood biological paradigm, namely ADC as a response biomarker following high dose RT.

Next, we demonstrated that LPM could appropriately describe ADC distributions of varying complexity, across two untreated xenograft models, with multiple model components being determined to account for modes of tumor heterogeneity. We then showed three important advantages of applying LPM to analyze the ADC data, all of which would not be possible using conventional image analysis methods. 

\textbf{Firstly}, in providing method technical validation, through a leave-one-out approach, we showed that it was possible to detect outliers in control groups. It is common to have variation in control group imaging biomarker values and this can substantially limit the ability of any biomarker to detect biological differences between small cohorts of control and treated animals (\cite{deJongM}). In the era of personalized medicine that employs tumor models of increasing biological relevance and complexity (\cite{SharplessNE}), the ability to exclude atypical tumors from cohort-wise analysis is of increasing importance. LPM enables outliers to be identified and excluded based on robust statistical methods. 

\textbf{Secondly}, any pair (pre- and post-) of ADC values can be assigned a probability (p-value or Z-score) that they are associated with variation observed within the control group, or are statistically different and thus can be considered belonging to a treatment group. By calculating the volume of voxels in each category, LPM quantifies the minimal amount of responding tissue (i.e. a lower bound) that can be detected; more voxels may respond, but cannot be distinguished from non-responding voxels within the distribution overlapping with control. Here all tumors showed some response, but the range of the lower bound on responding volumes varied by approximately 2.5 fold in LoVo and approximately four-fold in HCT116. 

\textbf{Thirdly}, this feature enables response detection on a sample by sample basis, without the need for spatial mapping, e.g. image segmentation and pre- post- treatment coregistration. This is possible since LPM models variation within control data and then can account for this in the treatment group, identifying the number of voxels that are different within the frequency distribution of data, as opposed to the spatial distribution. The key finding of this study was that LPM is substantially more powerful than conventional cohort-based statistical methods for analysing imaging data. Indeed, approximately 16-18 times as much data from conventional analyses (size and mean ADC) would be required to detect changes with equivalent power compared to an LPM analysis, equating to a 4 fold increase in power. 

The implications of these data are substantial. Once a control model is established, the need for similar animal numbers in the treatment group is diminished considerably. Subsequent studies for a known animal model would require a small number of new control animals (to establish equivalence with banked control data). Then very small cohorts can be tested for a given therapy. In particular, LPM can identify response on a per tumor basis with greater significance than seen in a conventional t-test analysis of control versus treatment cohorts. This would allow reduction in animal numbers, with welfare benefits (\cite{WorkmanP2}), and the ability to identify individual responders in small studies of therapies where different tumors with varying biology are treated. This may be attractive for avatar studies where patient derived samples are used to generate PDx and CDx models (\cite{MalaneyP}) and in co-clinical trials where multiple therapies are tested against animal models with different genetic knockdown/knockout features (\cite{ClohessyJG}). 

The automatic process of building an individual LPM model and computing its errors takes less than 5 minutes on typical hardware. However, the process must be performed multiple times during model selection and validation. The model selection process for LoVo required 10 models to be constructed, whereas HCT116 required 11. Leave-one-out validation required an additional 8 models for LoVo and 13 for HCT116, representing each possible leave-one-out control combination. Total run-time was less than 4 hours, making it feasible to perform multiple complete analyses per day.

The LPM method described here has some limitations. As the volume of responding tissue is computed by excluding all variation which cannot be interpreted as normal control development, this value is strictly a lower bound. This bound however, is appropriate for use as part of the null hypothesis test. Our method determines this estimate without labelling individual voxels of data, but instead operates by fitting the entire data ADC distributions, learning the correlations between those from two time points. In so doing LPM can estimate the volume of treatment response without having to solve the ill-posed problem of voxel to voxel registration - where investigators attempt to produce one-to-one mapping between voxels from images at different time points in tumors that change in shape and volume over time (\cite{OConnorJP2}). This does however prevent LPM in its current form generating voxel level treatment response maps, which might otherwise be assumed possible for a method which estimates volume of treatment response.

If the control cohort is not sufficient to describe control variation then treatment volume can be overestimated by inappropriately attributing previously unseen control variation to treatment. This is the same problem as missing high sources of control variation when applying a conventional t-test, but with the problem multiplied for a higher dimensional model. Translation of the technique requires further technical and biological validation, though showing consistency in results across multiple models and therapies, with data from different laboratories (\cite{DoblasS}). Clinical application may also be possible, with collection of the necessary data in an appropriate control group.

The method is protected from model construction problems that avoid over-interpretation of results. For instance, a highly atypical example will have a correspondingly high $\chi^2_D$, and since quantity error covariances are scaled by this value, the statistical significance of treatment estimates is penalized for poorly modelled data. Large quantity errors can generally be attributed to poor models, for example with few control data sets, but this problem can be reduced by adding additional (valid) training data. Equally, if contamination in the form of outliers is included in control data, the additional variability introduced in the control model reduces the ability to measure treatment, again penalising the statistical significance of results. While this reduces the statistical power of the method, it increases robustness by providing a working analysis which gives a valid, yet more limited, lower bound on volume changes.

\section{Conclusion}
\label{sec:conclusions}

We have shown that LPM can remove unwanted biological variation in image data (from growth) for tumors of varying spatial heterogeneity. This substantially increases sensitivity to treatment-induced change, thus increasing statistical power. Once control models are constructed, LPM enables significant changes to be detected for single tumors. This has important implications for 3Rs (specifically reduction in animals) and LPM may facilitate design of complex preclinical avatar and co-clinical trial experiments by providing adequate power to small cohort sizes.

\section*{Acknowledgements}

We woud like to thank Isabel Peset, Francesca Trapani, Garry Ashton, Caron Abbey and Steve Bagley of the Cancer Research UK Manchester Institute, University of Manchester for additional support.


\section*{Funding}

1. The Leverhulme Trust funding (grant RPG-2014-019) to N.A. Thacker 
2. Royal College of Radiologists Small Project Grant to J.P.B. O'Connor
3. Cancer Research UK (CRUK) Clinician Scientist award (grant C19221/A15267) to J.P B. O'Connor; 
4. CRUK and EPSRC Cancer Imaging Centre in Cambridge and Manchester funding to The University of Manchester (grant C8742/A18097) to N.A. Thacker, K.J. Williams and J.P.B. O'Connor



\bibliographystyle{abbrv}
%
%

\bibliography{Document}



\newpage
\begin{center}
\pagestyle{empty}
\includegraphics[scale = 0.98]{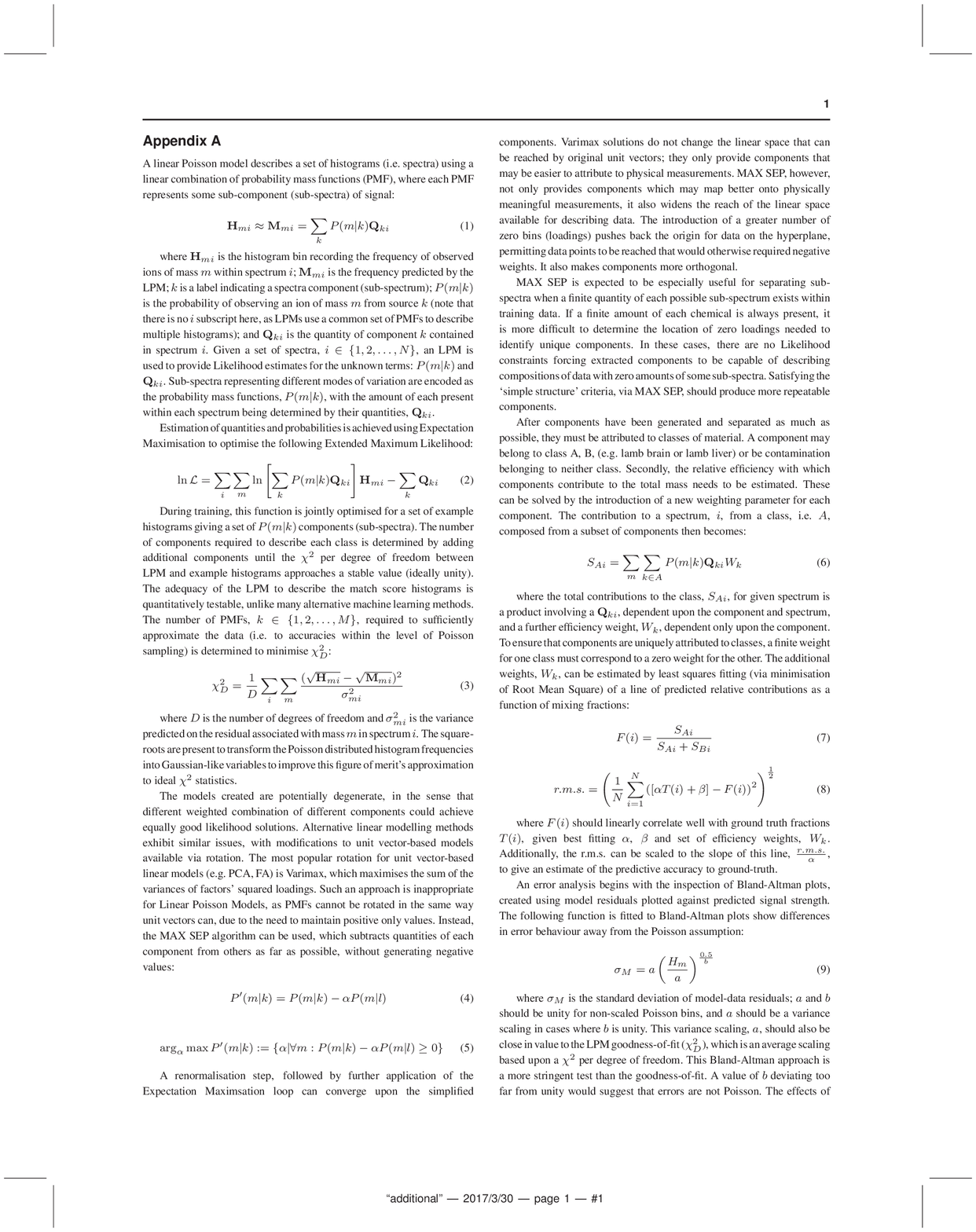}
\newpage
\includegraphics[scale = 0.98]{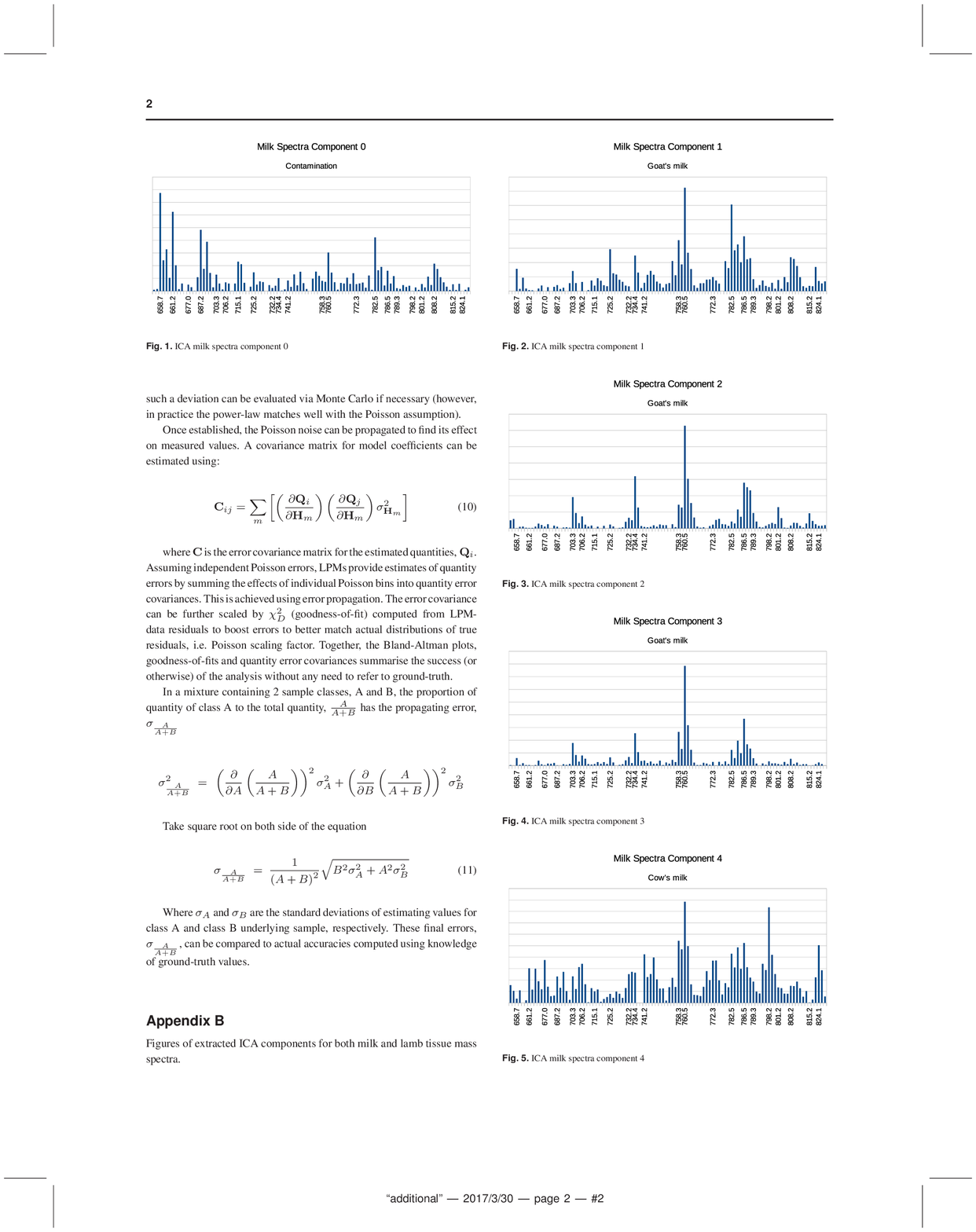}
\newpage
\includegraphics[scale = 0.98]{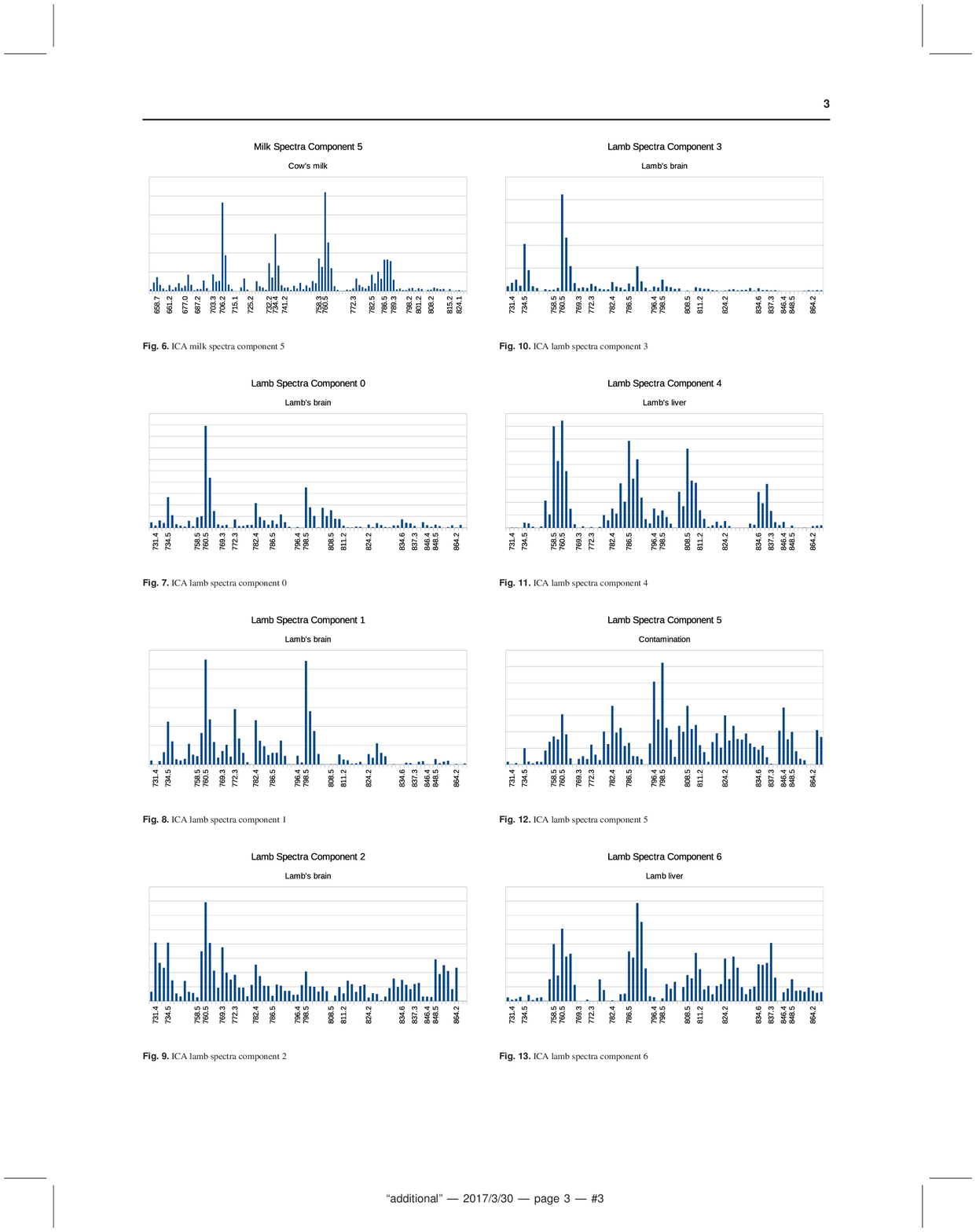}
\newpage
\includegraphics[scale = 0.98]{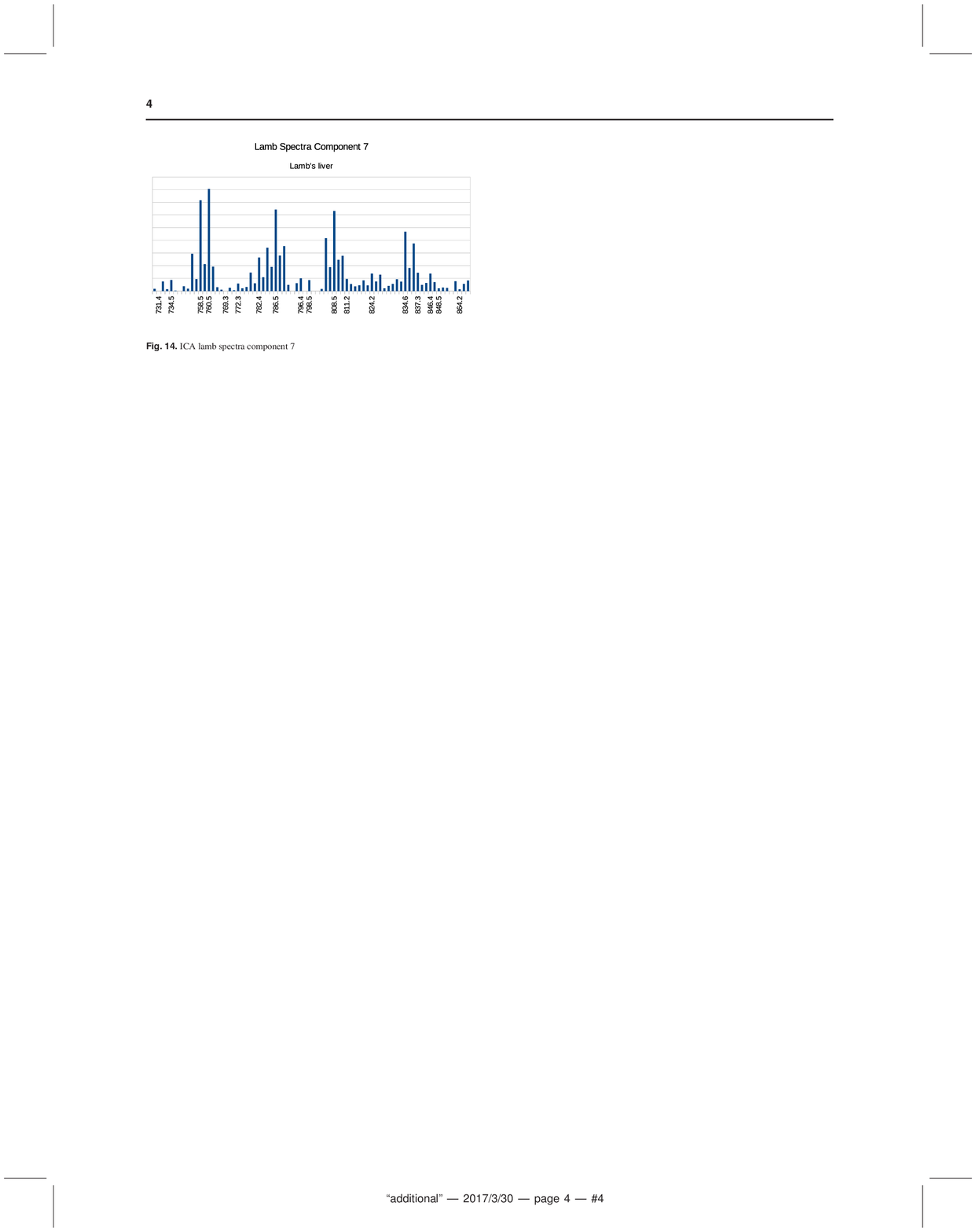}
\end{center}

\end{document}